\documentclass[10pt,conference]{IEEEtran}

\usepackage{cite}
\usepackage[table,xcdraw]{xcolor} %cor na tabela

\ifCLASSINFOpdf
   \usepackage[pdftex]{graphicx}
   \graphicspath{{figs/}}
   \DeclareGraphicsExtensions{.pdf,.jpeg,.png}
\else
   \usepackage[dvips]{graphicx}
   \graphicspath{{../figs/}}
   \DeclareGraphicsExtensions{.eps}
\fi
\usepackage[cmex10]{amsmath}

\usepackage{amsthm}

\usepackage{algorithmic}

\usepackage{array}

\ifCLASSOPTIONcompsoc
  \usepackage[caption=false,font=normalsize,labelfont=sf,textfont=sf]{subfig}
\else
  \usepackage[caption=false,font=footnotesize]{subfig}
\fi

\usepackage{url}
\makeatletter
\def\ps@IEEEtitlepagestyle{%
  \def\@oddfoot{\mycopyrightnotice}%
  \def\@oddhead{\hbox{}\@IEEEheaderstyle\leftmark\hfil\thepage}\relax
  \def\@evenhead{\@IEEEheaderstyle\thepage\hfil\leftmark\hbox{}}\relax
  \def\@evenfoot{}%
}
\def\mycopyrightnotice{%
  \begin{minipage}{\textwidth}
  \centering \scriptsize
  Copyright~\copyright~2020 IEEE. Personal use of this material is permitted. Permission from IEEE must be obtained for all other uses, in any current or future media, including reprinting/republishing this material for advertising or promotional purposes, creating new collective works, for resale or redistribution to servers or lists, or reuse of any copyrighted component of this work in other works. The original IEEE publication can be accessed at https://ieeexplore.ieee.org/document/9265999
  \end{minipage}
}
\makeatother

\begin{document}

\title{MyFood: A Food Segmentation and Classification System to Aid Nutritional Monitoring}

%------------------------------------------------------------------------- 
% change the % on next lines to produce the final camera-ready version 
\newif\iffinal
% \finalfalse
\finaltrue
\newcommand{\cmtid}{63}
%------------------------------------------------------------------------- 

% author names and affiliations
% use a multiple column layout for up to two different
% affiliations

\iffinal

% author names and affiliations
% use a multiple column layout for up to three different
% affiliations
\author{\IEEEauthorblockN{Charles N. C. Freitas}
\IEEEauthorblockA{Department of Computing \\ Universidade Federal Rural de \\ Pernambuco, Recife, PE \\
Email: charles.freitas@ufrpe.br}
\and
\IEEEauthorblockN{Filipe R. Cordeiro}
\IEEEauthorblockA{Department of Computing \\ Universidade Federal Rural de \\ Pernambuco, Recife, PE \\
Email: filipe.rolim@ufrpe.br}
\and
\IEEEauthorblockN{Valmir Macario}
\IEEEauthorblockA{Department of Computing \\ Universidade Federal Rural de \\ Pernambuco, Recife, PE \\
Email: valmir.macario@ufrpe.br}}

% conference papers do not typically use \thanks and this command
% is locked out in conference mode. If really needed, such as for
% the acknowledgment of grants, issue a \IEEEoverridecommandlockouts
% after \documentclass

% for over three affiliations, or if they all won't fit within the width
% of the page, use this alternative format:
% 
%\author{\IEEEauthorblockN{Michael Shell\IEEEauthorrefmark{1},
%Homer Simpson\IEEEauthorrefmark{2},
%James Kirk\IEEEauthorrefmark{3}, 
%Montgomery Scott\IEEEauthorrefmark{3} and
%Eldon Tyrell\IEEEauthorrefmark{4}}
%\IEEEauthorblockA{\IEEEauthorrefmark{1}School of Electrical and Computer Engineering\\
%Georgia Institute of Technology,
%Atlanta, Georgia 30332--0250\\ Email: see http://www.michaelshell.org/contact.html}
%\IEEEauthorblockA{\IEEEauthorrefmark{2}Twentieth Century Fox, Springfield, USA\\
%Email: homer@thesimpsons.com}
%\IEEEauthorblockA{\IEEEauthorrefmark{3}Starfleet Academy, San Francisco, California 96678-2391\\
%Telephone: (800) 555--1212, Fax: (888) 555--1212}
%\IEEEauthorblockA{\IEEEauthorrefmark{4}Tyrell Inc., 123 Replicant Street, Los Angeles, California 90210--4321}}

\else
  \author{Sibgrapi paper ID: \cmtid \\ }
\fi

% make the title area
\maketitle

% As a general rule, do not put math, special symbols or citations
% in the abstract
\begin{abstract}
The absence of food monitoring has contributed significantly to the increase in the population's weight. Due to the lack of time and busy routines, most people do not control and record what is consumed in their diet. Some solutions have been proposed in computer vision to recognize food images, but few are specialized in nutritional monitoring. This work presents the development of an intelligent system that classifies and segments food presented in images to help the automatic monitoring of user diet and nutritional intake. This work shows a comparative study of state-of-the-art methods for image classification and segmentation, applied to food recognition. In our methodology, we compare the FCN, ENet, SegNet, DeepLabV3+, and Mask RCNN algorithms. We build a dataset composed of the most consumed Brazilian food types, containing nine classes and a total of 1250 images. The models were evaluated using the following metrics: Intersection over Union, Sensitivity, Specificity, Balanced Precision, and Positive Predefined Value. We also propose an system integrated into a mobile application that automatically recognizes and estimates the nutrients in a meal, assisting people with better nutritional monitoring. The proposed solution showed better results than the existing ones in the market. The dataset is publicly available at the following link http://doi.org/10.5281/zenodo.4041488.

\end{abstract}

% no keywords

% For peerreview papers, this IEEEtran command inserts a page break and
% creates the second title. It will be ignored for other modes.
\IEEEpeerreviewmaketitle

\section{Introduction}

Poor diet and lack of physical activities are among the main factors that contribute to the population increase in weight \cite{CIC11}. This is usually related to intensive routines and a lack of knowledge about a healthy diet. These aspects have a high impact on world health because, with the increase in the population's weight, the number of people with chronic non-transmissible diseases (e.g., diabetes and hypertension) has grown \cite{WOF20}.

According to the World Health Organization (WHO), obesity is more common than innutrition \cite{WHO20}, i.e., the population is becoming increasingly overweight and less nourished. In Brazil, the obesity rate has increased by 67.8\% in the last thirteen years, according to the Health Ministry \cite{Saude19}. Although obesity rates have grown in recent years, the survey in \cite{Saude19}  states that this index has been established since 2015, corresponding today to around 18.9\% of the entire population.

Despite the current scenario, it has been seen an increasing interest of people towards a healthier lifestyle and nutritional information for diet \cite{Steingoltz18}. % Faced with this scenario, people have increasingly sought to carry out diets and worry about health and well-being, in search of finding relationships about dietary patterns
% healthier \cite{FIESP18}. 
However, any diet must be accompanied by a nutritionist, which will require monitoring the number of calories and nutrients in the meals of each person. For this purpose, several intelligent systems have been proposed to help people maintain the right level of food consumption and awareness of their diet by monitoring their eating habits \cite{Saude19}.
%However, this can be compromised over time, with the need for more practical and effective food monitoring \cite{Hassannejad17}.

% The increasingly recurrent use of intelligent systems has helped to improve nutritional adequacy \cite{Freitas18}. Currently, technologies have helped people to maintain a good level of food consumption and awareness of their diet by monitoring their eating habits \cite{Saude19}. In recent years, much work and research has shown that machine learning and computer vision techniques have a high potential in building automatic food recognition systems to estimate the nutritional values present at each meal.

Current nutritional monitoring apps use traditional methods for food recognition, with algorithms based on classification, using convolutional neural networks (CNN) \cite{Kagaya14,Liu16, Temdee17, Pandey17}. 
However, such applications are either semi-automatic, which identifies a group of possible types of food and requires user interaction or have incomplete information, such as the amount of food and calorie information. In this work, we propose the use of state-of-the-art (SOTA) segmentation algorithms to segment foods consumed in Brazilian meals, providing nutritional information to help the monitoring of diet.

% to identify t The segmentation technique applied in this work, on the other hand, brings a more specialized classification approach in regions of interest in the image, which when inserted in the food context brings benefits related to nutritional but assertive monitoring about the foods contained in the image.

For food recognition problems, it is common to use data sets such as Food101 \cite{food101}, composed of 101 food categories, with 101,000 images. However Food101 does not have images with various food classes, making it difficult to use in real-world settings. In this context, we chose to build our own image data set to meet local needs, with the most consumed food in Brazil. The built data set will be publicly available.

% However, none of the existing apps provides segmen On the other hand, image segmentation simplifies the image representation process and provides contextual information that can be used for the classification \cite{Carlsson18}. In this regard, segmentation has a more specialized classification approach in regions of interest in the image, which when inserted in the food context can bring benefits related to the estimation of the volume of food, providing nutritional but assertive monitoring in relation to the foods contained in the food image.

% There are works which have used image segmentation of food images for nutritional monitoring \textcolor{red}{[??,??]}. Thus, 

This works evaluates SOTA methods of segmentation and classification applied to a Brazilian food dataset, based on the most consumed types of dietary. We also build a prototype of an app to use with the best methods evaluated, which showed improvements compared to existing apps. 

Our main contributions are described below:

\begin{itemize}
    \item{Comparison of state-of-the-art approaches for segmentation of Brazilian type of food.}
    \item{A dataset with 1250 images and 9 classes of food (with annotations of the images)}
% 	\item Proposed solution for nutritional monitoring with Brazilian foods, with food recognition in a more specialized way, with application of the segmentation method.
    \item {Mobile application with integrated food segmentation and classification system to aid nutritional monitoring, which shows better results than existing solutions in the market.}
\end{itemize}

% SIBGRAPI 2020 \cite{Sibgrapi2020}.

\section{Related Work}

% 

% According to Guo et al. \cite{Guo17}, performing segmentation without knowing the exact identity of all objects in the scene is an important part of our visual understanding process, which can give us a powerful model to understand the world and also be used to improve or enhance existing computer vision techniques\cite{Guo17}.

Deep learning has recently been used to solve complex problems, such as image segmentation and classification \cite{Walsh19}. Recent approaches in literature have been applied for food recognition in images. Zhang et al. \cite{Zhang15} proposed a food image recognition system, using CNN, which can recognize about 100 types of food. In \cite{Sun19}, Sun et al. propose a mobile application for recognition of food items from a single image. They used the CNN architecture based on Mobilenet \cite{Howard2017MobileNetsEC} with the detection mechanism YOLO\_v2 \cite{redmon2016yolo9000} to generate bounding boxes for each class of food. A performance of 76.36\% of average precision was obtained in UECFood100 and 75.05\% in UECFood256.

% Food segmentation is a promising area, since it is intended to have a more specialized recognition of the objects that make up the image, facilitating the process of nutritional estimation and analysis of people's eating habits for health care. 
Shimoda et al. \cite{Shimoda15} proposed a new regional targeting method that combines the ideas from the RCNN \cite{Girshick13} network. The regions proposed in the method are generated by selective search, extracting activation resources from the network and applying a support vector machine (SVM) to evaluate the proposals and produce bounding boxes for the objects. The authors adjusted the pre-trained network to the ImageNet \cite{imagenet_cvpr09} database using the PASCAL VOC \cite{pascal-voc-2012} data set with 20 categories. The result for the metric intersection over union (IoU) was greater than 50\% between the detected bounding box and the actual bounding box, outperforming or RCNN in relation to the food detection region, and also a PASCAL VOC detection task.

% Such works bring grounded reflections, addressing the challenges of recognizing food in images, with the aim of enhancing the use of technologies in favor of strengthening healthy eating habits. 

In this research, five deep learning architectures commonly used for image segmentation tasks were evaluated: FCN \cite{Long15}, Segnet \cite{Badrinarayanan17}, ENet \cite{Paszke16}, DeepLabV3+ \cite{deeplabv3plus2018} and Mask RCNN \cite{He18}. These models are SOTA in the segmentation of image task and they were evaluated with the built dataset.

\section{Methodology}

For this work, Python was used with the library Keras  \cite{chollet2015keras} which provides a simple and objective way to create a variety of deep learning models using TensorFlow. The models implemented in this project were processed using the Google Colaboratory tool (Google Colab) \cite{colab20}, which allows the use of the Jupyter Notebook service hosted on Google servers, with free access to computational resources, including 16GB Tesla P100 GPU and 20GB of RAM.

\subsection{Construction of the Brazilian Food Image Dataset}

% A survey carried out by the brazilian platform \textit{Vigilantes do Peso}  \cite{Boaforma18} ranked the 50 most consumed foods by the Brazilian population. Based on this survey, we filtered the 10 most consumed foods on a daily basis. The survey was created on the Google® forms platform and shared on social media, with the aim of acquiring the necessary information for constituting the image base. 59 responses were obtained, and the 10 foods were ranked as shown in the 

\begin{figure}[h]
    \centering
    \includegraphics[width=3.5in]{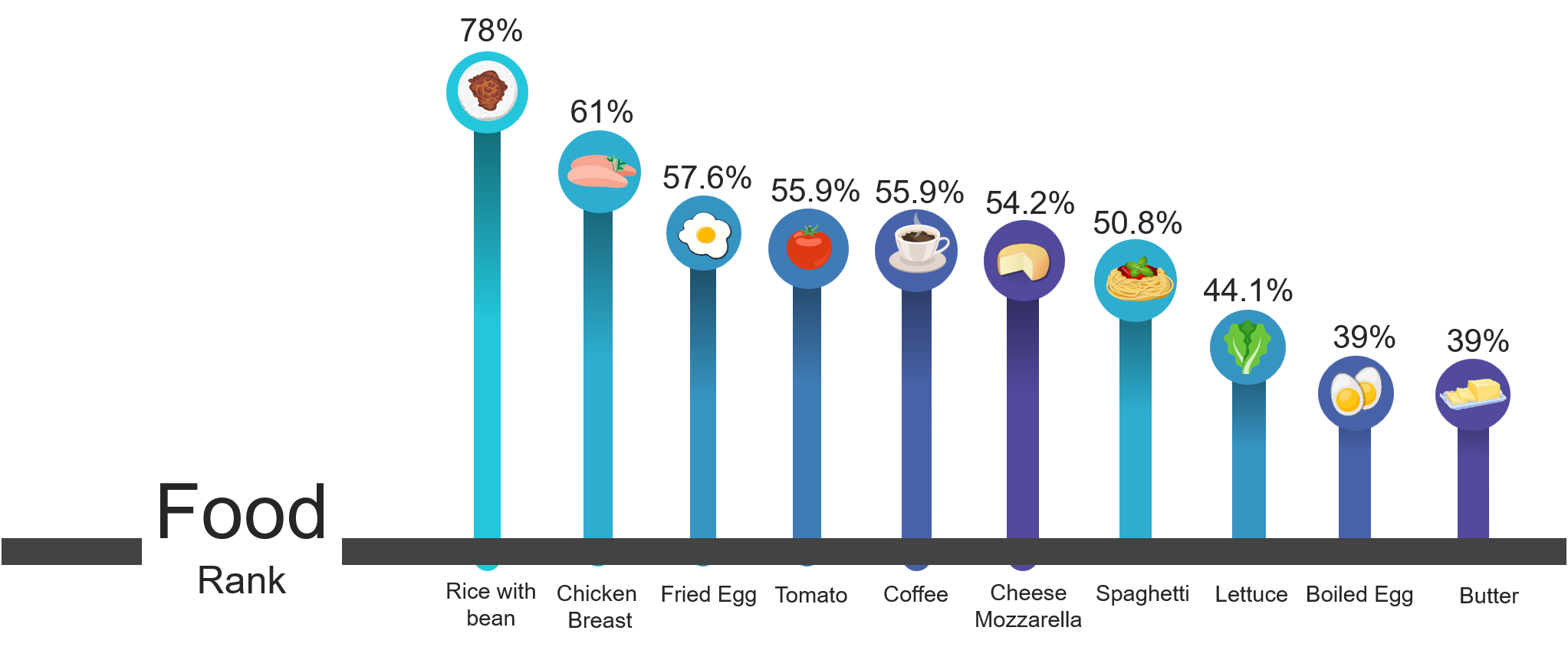}
    \caption{Results of the survey about the 10 most consumed foods.}
    \label{rank_10_foods}
\end{figure}

The image dataset of this work was built  through a survey carried out by the Vigilantes do Peso \cite{Boaforma18} platform, which gathered the classification of the 50 most consumed foods by the Brazilian population. The amount of 50 foods was restricted based on a research and observational study to filter the 10 foods most consumed by people on a daily basis. A list of the most consumed foods is shown in Figure~\ref{rank_10_foods}.

From the results obtained, it was possible to determine 9 classes of foods that were used in the research experiments. The selected classes were \textit{rice, beans, boiled egg, fried egg, pasta, salad, roasted meat, apple}, and \textit{chicken breast}. To optimize the image acquisition process, the tomato and lettuce foods were grouped into salads, and the coffee, butter, and mozzarella cheese foods were removed due to the complexity of acquiring the images, being replaced by apples and roasted meat. In the end, we obtained an average of 125 images per class and a total of 1250 images to conduct the evaluation experiments, using 60\% of dataset for training, 20\% for validation and 20\% for test.

\begin{figure}[h]
    \centering
    \includegraphics[width=3.5in]{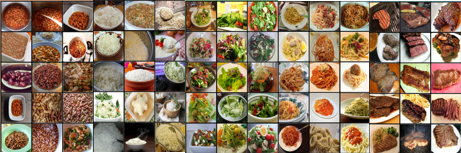}
    \caption{Images from the Brazilian food dataset.}
    \label{dataset}
\end{figure}

To obtain the images corresponding to the defined classes, a search and image capture script was developed on the servers of Google Image, Flickr and the Bing API, in order to facilitate the process of acquiring food images. All images were maintained according to their characteristics and were only resized to the 512x512 pixel standard. In the Figure ~\ref{dataset} a section of our database is shown. Our dataset can be downloaded in Zenodo platatform \cite{freitas_charles_2020_4041488} at http://doi.org/10.5281/zenodo.4041488.

\begin{figure}[h]
    \centering
    \includegraphics[width=1.5in]{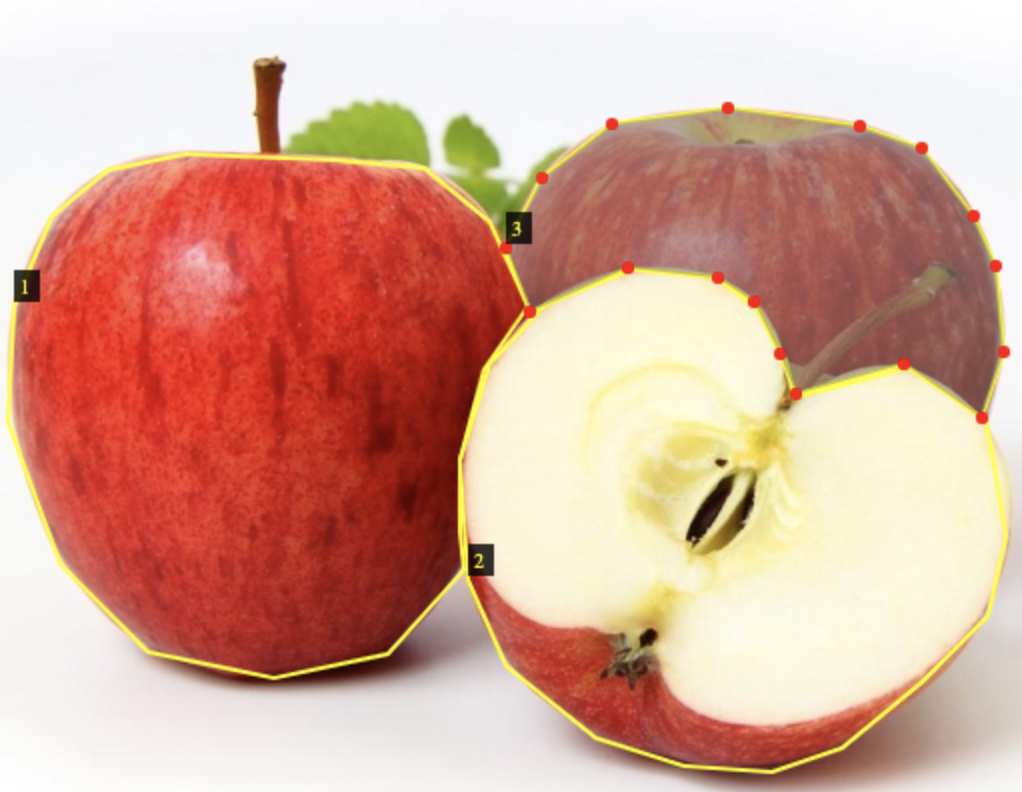}
    \caption{Image annotation with VIA.}
    \label{sample_anotation}
\end{figure}

In each image of the training, validation and test subset, the process of labeling the food segmentation masks was performed, as shown in the figure ~\ref{sample_anotation}. The process was performed manually with the support of the VGG Image Annotator (VIA) \cite{dutta2016via} tool, which saves the notes in a JSON file and each mask is represented by a set of polygon points. The masks are defined according to the characteristics (contours and shapes) of each image, assigning the segmentation masks corresponding to each identified class.

% Figure~\ref{sample_anotation} shows an example of annotating the image using the VGG Image Annotator (VIA)\cite{dutta2016via}.

\subsection{Compared Approaches}

% The image segmentation models implemented are based on a encoder-decoder architectures. 
We compare the following segmentation methods from SOTA for food segmentation:

\textbf{FCN}\cite{Long15}: Fully Convolutive Networks (FCN) is built from locally connected layers \cite{Long15}, using a convolutional neural network to transform image pixels into categories of pixel \cite{Smola19}. J. Long et al. \cite{Long15} were the first to develop a fully convolutive network trained from end to end for image segmentation. The authors modified known architectures (AlexNet, VGG16, GoogLeNet) to have an arbitrarily sized entry, replacing all fully connected layers with convolutional layers \cite{Ouaknine18}. In this research, the FCN8 model was used, in which the VGG16  network is duplicated and the final classifying layer is discarded, converting all fully connected layers into convolutions.

\textbf{Segnet}\cite{Badrinarayanan17}: SegNet is a deep encoder-decoder architecture for segmenting pixel-class of various classes, developed by members of the Vision Group and Computer Robotics at the University of Cambridge, United Kingdom. SegNet was mainly motivated by scene understanding applications. This research used a simplified version of SegNet, called SegNet-Basic \cite{Badrinarayanan17}, which has four encoders and four decoders, which allows us to explore many different variants (decoders) and train them in a reasonable time.

\textbf{ENet} \cite{Paszke16}: ENet's architecture is largely based on the ResNet network \cite{Kaiming15}, being summarized to a main structure and several branches that separate, but also merge again through adding elements. In \cite{Paszke16}, the authors propose a new neural network architecture optimized for fast inference and high precision, with a focus use on mobile devices.

\textbf{DeepLabV3+}\cite{deeplabv3plus2018}: DeepLab is a  targeting model, designed by the Google research group in 2016 \cite{deeplabv3plus2018}. The DeepLabv3+ proposal, according to the authors, is to extend the predecessor model (DeepLabv3) by applying the concept of atrous separable convolution composed of a deep convolution and a clockwise convolution (1$\times$1 convolution with depth as input). Also employing an encoder-decoder structure, in which the encoder module processes contextual information at various scales applying the atrous convolution, while the simple yet effective decoder refines the segmentation results along with the limits of the object \cite{deeplabv3plus2018}.

\textbf{Mask RCNN}\cite{He18}: Mask RCNN \cite{He18} is an extension of Faster RCNN, which is used for detection tasks of objects \cite{SHARMA192}. The Mask RCNN was developed by the Facebook research group and published in 2018, with the proposal of instance segmentation, in which the task of identifying each instance of the object the pixels corresponding to the classes contained in the image \cite{He18}. In summary, we can say that Mask RCNN combines the two main networks - Faster RCNN and FCN in a big architecture, whose loss function for the model is the total loss of the classification \cite{Dwivedi19}.

All the architectures implemented in the research were based on projects hosted in open source repositories. We made adjustments to the hyperparameters and the dataset loader to better adapt the research's analysis and experimentation.

\subsection{Parameters}

The parameters used in this research are based on the original papers of the architectures. Adjustments were applied to some parameters, such as input size, learning rate, decay, and epochs, to adapt the models to our dataset. We trained the models using momentum value of 0.9, image sizes of 224 $\times$ 224 pixels, 100 epochs of traning, batch sizes of 2, 10 and 32, and learning rate variations between $10^{- 3}$, $10^{- 4}$ and $5^{- 4}$, according to the basic definition of each architecture. Table ~\ref{table_params_comparative} shows the best parameters found for each algorithm.

\begin{table}[h]
\centering
\footnotesize 
\caption{Parameters used for each model.}
\label{table_params_comparative}
\begin{tabular}{|c|c|c|c|c|c|}
\hline
\rowcolor[HTML]{000000} 
{\color[HTML]{FFFFFF} \textbf{Model}} & {\color[HTML]{FFFFFF} \textbf{Optimizer}} & {\color[HTML]{FFFFFF} \textbf{Rate}} & {\color[HTML]{FFFFFF} \textbf{Decay}} & {\color[HTML]{FFFFFF} \textbf{\begin{tabular}[c]{@{}c@{}}Batch\\ Size\end{tabular}}} & {\color[HTML]{FFFFFF} \textbf{Backbone}} \\ \hline
FCN                                   & SGD                                       & 1E-2                                 & -                                     & 32                                                                                   & VGG16                                    \\ \hline
SegNet                                & SGD                                       & 1E-2                                 & -                                     & 32                                                                                   & -                                        \\ \hline
ENet                                  & Adam                                      & 5E-4                                 & -                                     & 10                                                                                   & -                                        \\ \hline
DeepLabV3+                            & SGD                                       & 1E-2                                 & -                                     & 32                                                                                   & MobileNet                                \\ \hline
Mask RCNN                             & SGD                                       & 1E-3                                 & 1E-4                                  & 2                                                                                    & Resnet101                                \\ \hline
\end{tabular}
\end{table}

\subsection{Metrics}
\label{sec:metrics}

We evaluated the performance of each model according to the metrics used in literature: Intersection over Union (IoU), Positive Prediction Value (PPV), Sensitivity (SE), Specificity (SP), and Balanced Accuracy (BAC).

The metrics used are based on the values of true positive (TP), false positive (FP), true negative (TN) and false negative (FN). These values are calculated based on the binary segmentation ouput of each method. Each of the metrics are described bellow.

% \begin{ceqn}
% \begin{align} 
% \label{iou}
% 	IoU = \frac{| A \cap B |}{| A \cup B |} = \frac{VP}{VP + FN +FP}
% \end{align}
% \end{ceqn}
The Intersection over Union (IoU), also known as the Jaccard index, is a simple and extremely effective rating metric \cite{Tiu19}, which calculates the area of overlap between the predicted segmentation ($A$) and the ground truth (GT) ($B$), divided by the union area between the predicted segmentation and the GT segmentation, as shown in the equation ~\ref{iou}.
\begin{equation}
\label{iou}
    IoU = \frac{| A \cap B |}{| A \cup B |} = \frac{TP}{TP + FN + FP}
\end{equation}

Positive prediction is calculated on all labels by measuring the number of times that a given class has been classified correctly. The prediction determined by the label considers only one class and measures the number of times that a specific label was correctly predicted (TP) normalized by the total number of positive predictions, as demonstrated in Equation ~\ref{ppv}.

\begin{equation}
\label{ppv}
    PPV = \frac{TP}{TP + FP}
\end{equation}

Sensitivity (SE) is calculated as the number of correct positive predictions divided by the total number of positives. It is also called recall, it represents the effectiveness of the algorithm in correctly classifying the pixels of the image object \cite{Andrade17}. Sensitivity is described according to the equation ~\ref{SE}. 

\begin{equation}
    \label{SE}
	SE = \frac{TP}{TP + FN}
\end{equation}

Specificity (SP) is calculated as the number of correct negative predictions divided by the total number of negatives. Specificity is calculated according to the equation ~\ref{SP}.
%Specificity measures the probability that the test will return negative to someone who does not have the characteristic, it represents the effectiveness of the algorithm in correctly classifying the background pixels of the image \cite{Andrade17}. 

\begin{equation}
    \label{SP}
	SP = \frac{TN}{TN + FP}
\end{equation}

Balanced Accuracy (BAC) is the average between specificity and sensitivity. The BAC is calculated according to the equation ~\ref{bac}.

\begin{equation}
    \label{bac}
	BAC = \frac{SE + SP}{2}
\end{equation}

\section{Results and Discussion}

% The experiments were run two times, with 50 and 100 epochs, seeking the best learning convergences of the models. 
We evaluated the methods FCN, Segnet, ENet and DeepLabV3+, according to the metrics described in section \ref{sec:metrics}. We run the model for 100 epochs, and the results are shown in Table~\ref{table_models}.

\begin{table}[h]
\centering
\footnotesize 
\caption{Segmentation results for the Brazilian food dataset.}
\label{table_models}
\begin{tabular}{|l|c|l|c|c|l|}
\hline
\rowcolor[HTML]{000000} 
\multicolumn{1}{|c|}{\cellcolor[HTML]{000000}{\color[HTML]{FFFFFF} \textbf{Model}}} & {\color[HTML]{FFFFFF} \textbf{IoU}} & \multicolumn{1}{c|}{\cellcolor[HTML]{000000}{\color[HTML]{FFFFFF} \textbf{SE}}} & {\color[HTML]{FFFFFF} \textbf{SP}} & {\color[HTML]{FFFFFF} \textbf{BAC}} & \multicolumn{1}{c|}{\cellcolor[HTML]{000000}{\color[HTML]{FFFFFF} PPV}} \\ \hline
FCN                                                                                 & \textbf{0.70(0.2)}                           & \textbf{0.81(0.2)}                                                                       & \textbf{0.99(0.02)}                         & \textbf{0.90(0.1)}                           & 0.79(0.2)                                                               \\ \hline
Segnet                                                                              & 0.52(0.2)                           & 0.64 (0.2)                                                                       & 0.97(0.05)                         & 0.81(0.1)                           & 0.69(0.2)                                                               \\ \hline
Enet                                                                                & 0.51(0.3)                           & 0.64(0.3)                                                                       & 0.98 (0.03)                         & 0.81 (0.1)                           & 0.69(0.2)                                                               \\ \hline
\begin{tabular}[c]{@{}l@{}}DeepLab\\ V3+\end{tabular}                               & 0.5(0.3)                            & 0.66 (0.3)                                                                       & 0.98(0.05)                         & 0.82(0.1)                           & 0.79(0,2)                                                               \\ \hline
\begin{tabular}[c]{@{}l@{}}Mask \\ RCNN\end{tabular}                                & \textbf{0.70(0.2)}                           & 0.76(0.2)                                                                       & 0.98(0.05)                         & 0.87(0.09)                          & \textbf{0.87(0.1)}                                                               \\ \hline
\end{tabular}
\end{table}

\begin{figure}[h]
    \centering
    \includegraphics[width=3.5in]{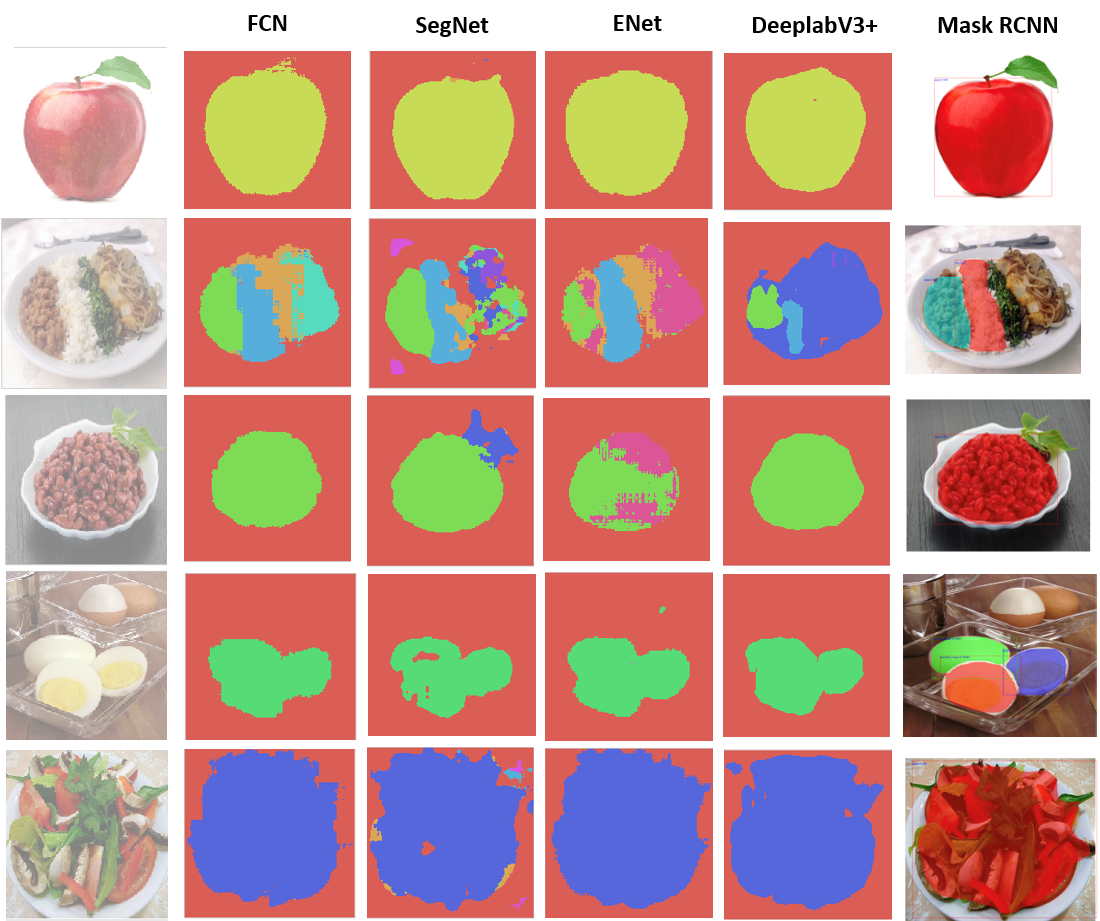}
    \caption{Segmentation of the models.}
    \label{predict_models}
\end{figure}

As shown in Table \ref{table_models}, the FCN model showed better results for most of the metrics. Although FCN used VGG16 backbone, it could generate more precise segmentation, with higher values of IoU and BAC. The Figure~\ref{predict_models} shows qualitative results of segmentation for each method, for five images of the data set.

From Figure~\ref{predict_models} we can see that most methods segment well the image when there is only one class of food in the image. However, when there are many classes, FCN segments better.

\subsection{Class analysis}

\begin{figure}[h]
    \centering
    \includegraphics[width=3.5in]{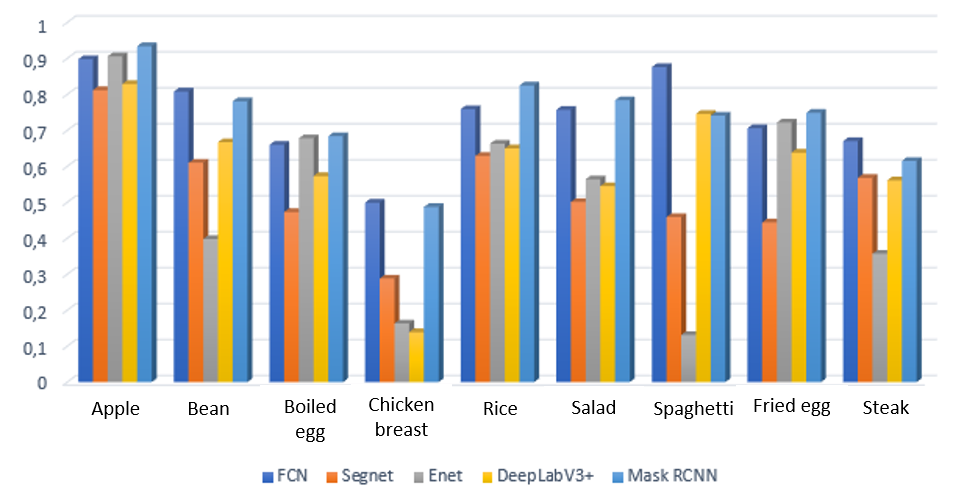}
    \caption{IoU values for each class.}
    \label{graficos_f}
\end{figure}

The analysis of IoU for each class is shown  in Figure~\ref{graficos_f}. It shows that the apple class was the class with the best results, meaning that it is the easiest one. The chicken breast class, on the other hand, was the one with the lowest rating among the models. The FCN and Mask RCNN models were the ones that obtained the best results in relation to most classes, with IoU values rates above 0.6.

\subsection{System architecture and prototype}

\begin{figure}[h]
    \centering
    \includegraphics[width=3.5in]{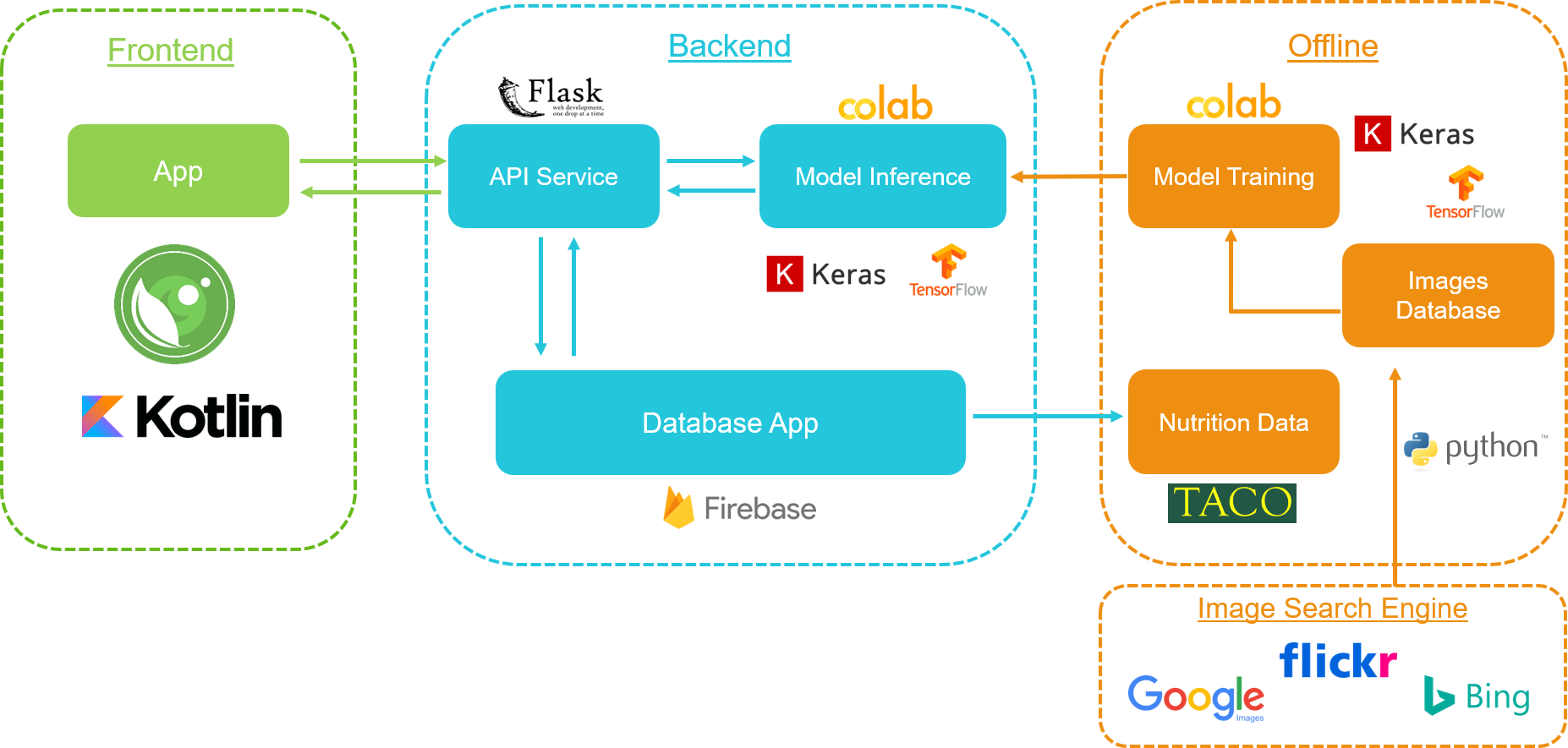}
    \caption{System Architecture of MyFood}
    \label{architecture_myfood}
\end{figure}

Our model was integrated with a mobile application, following the pipeline of capture, analysis, segmentation, and classification of food images, to create an application for food monitoring. The system consists of three parts: 1) \textit{offline}, which comprises the base of images and experiments carried out with the implemented models; 2) \textit{backend}, composed of the inference mechanisms (based on the model that obtained better results), a database with the nutritional information of the foods covered in the research and an API Service (Flask) to communicate with the application; and 3) a \textit{Frontend} composed by the application mobile that will be the user interaction interface with the system. In figure ~\ref{architecture_myfood} the complete architecture of the developed system is presented.

\begin{figure}[h]
    \centering
    \includegraphics[width=3.5in]{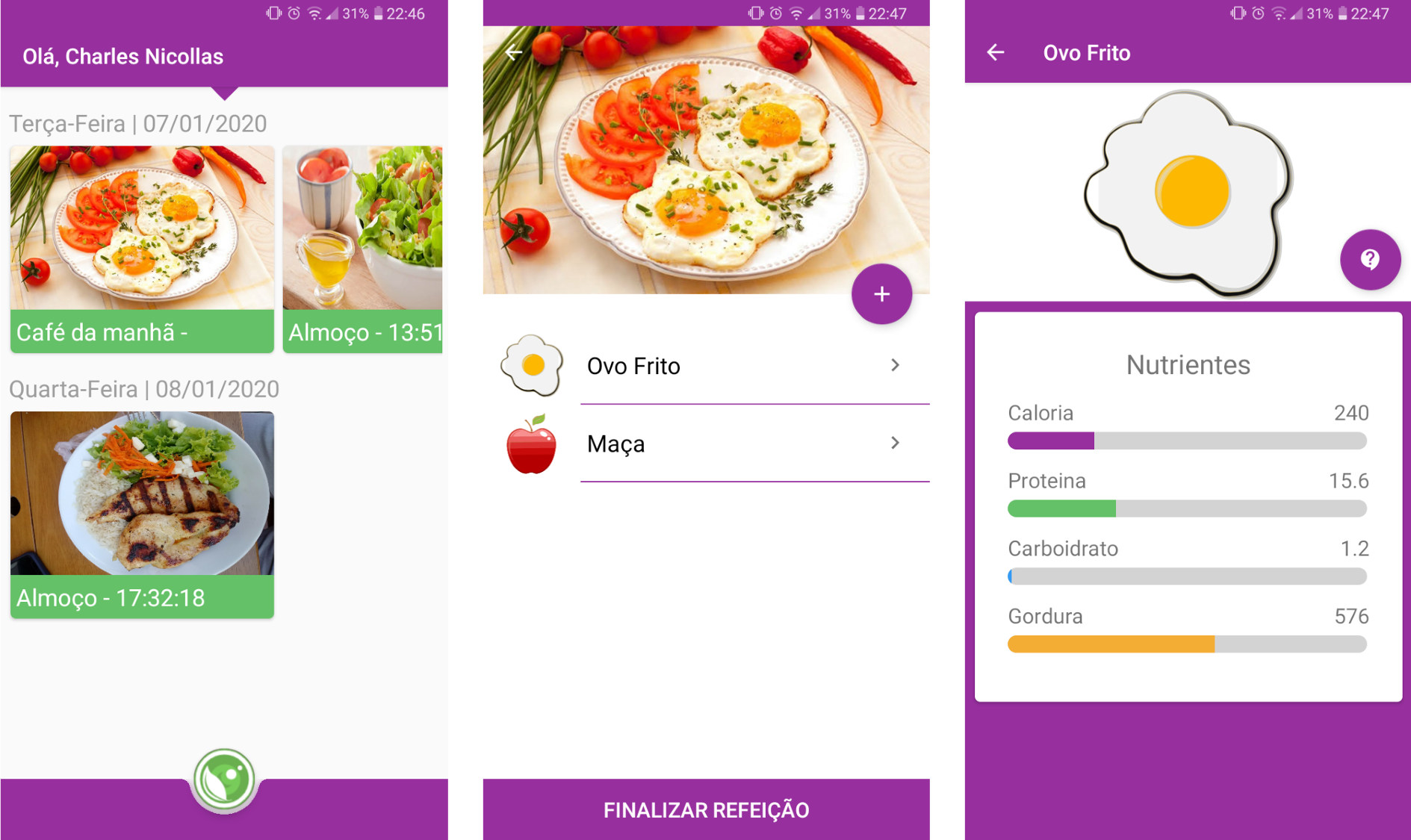}
    \caption{Prototype of the proposed app MyFood.}
    \label{prototipo}
\end{figure}

The model with the best results regarding the applied metrics was integrated into a mobile application, following the flow of capture, analysis, segmentation and classification of food images, with the purpose of creating an application for food monitoring. The prototype was developed following all current standards and methodologies for design and usability, to ensure greater acceptance and use of the application. Figure~\ref{prototipo} illustrates some screens of the prototype of the developed application.

\subsection{Comparison with existing apps}

Based on the shown results, the Mask RCNN and FCN models obtained the best results in relation to the applied metrics. Based on that, Mask RCNN was chosen for a comparative analysis with three apps from the market: Calorie Mama\cite{CalorieMama}, Lose It!\cite{LoseIt} and FoodVisor\cite{FoodVisor}). For this, 12 images of food  were taken using a mobile phone and we compared the results of segmentation and classification for each image. Figure~\ref{predict_mask} shows the results:

\begin{figure}[h]
    \centering
    \includegraphics[width=3.5in]{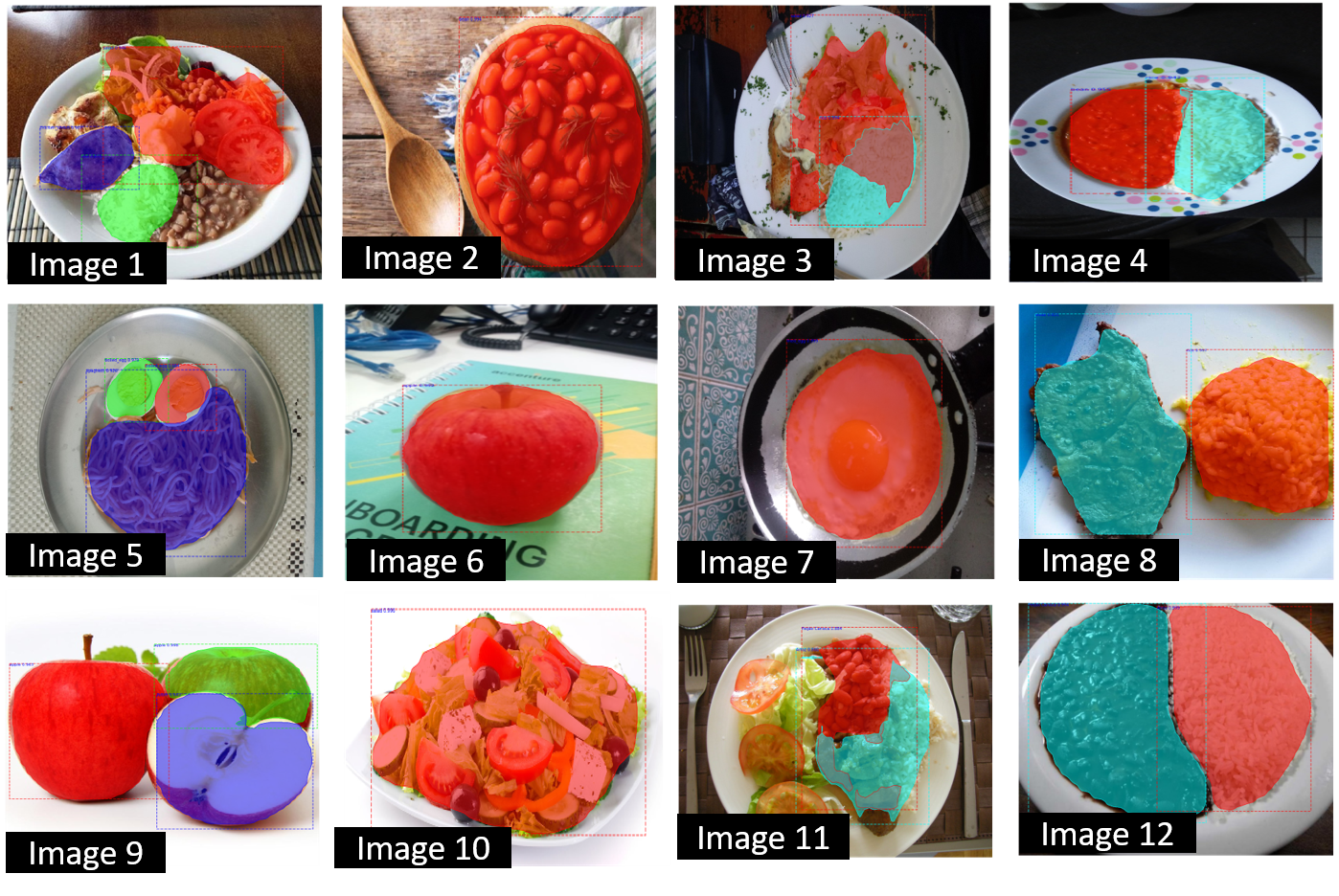}
    \caption{Segmentation results for the Mask RCNN model using our prototypal.}
    \label{predict_mask}
\end{figure}

Real images of meals and food were used in order to analyze our solution and other similar applications. For Myfood one of the high challenges is the identification of meals with multiple foods, in which it is necessary to process, classify and segment each item in the image. Table ~\ref{table_comparative_app} shows  the comparative analysis performed with each application, using the previous 12 images.

% Although the FCN model has a simpler architecture, and the basis for other models, 
% Mask RCNN was chosen because it uses in addition to FCN other concepts (RPN, RoI, etc.) that add to the potential for segmentation and classification of images.

\begin{table}[h]
\centering
\footnotesize 
\caption{Prediction analysis of using market apps and MyFood.}
\label{table_comparative_app}
\begin{tabular}{
>{\columncolor[HTML]{FFFFFF}}c |
>{\columncolor[HTML]{34FF34}}c |c|c|
>{\columncolor[HTML]{34FF34}}c |}
\cline{2-5}
{\color[HTML]{000000} \textbf{}}                                                       & \cellcolor[HTML]{000000}{\color[HTML]{FFFFFF} \textbf{Calorie Mama}} & \cellcolor[HTML]{000000}{\color[HTML]{FFFFFF} \textbf{Lose It!}} & \cellcolor[HTML]{000000}{\color[HTML]{FFFFFF} \textbf{FoodVisor}} & \cellcolor[HTML]{000000}{\color[HTML]{FFFFFF} \textbf{MyFood}} \\ \hline
\multicolumn{1}{|c|}{\cellcolor[HTML]{FFFFFF}{\color[HTML]{000000} \textbf{Image 1}}}  & \cellcolor[HTML]{FFC702}good                                         & \cellcolor[HTML]{FFC702}good                                     & \cellcolor[HTML]{FD6864}bad                                       & \cellcolor[HTML]{FFC702}good                                   \\ \hline
\multicolumn{1}{|c|}{\cellcolor[HTML]{FFFFFF}{\color[HTML]{000000} \textbf{Image 2}}}  & great                                                                & \cellcolor[HTML]{34FF34}great                                    & \cellcolor[HTML]{34FF34}great                                     & great                                                          \\ \hline
\multicolumn{1}{|c|}{\cellcolor[HTML]{FFFFFF}{\color[HTML]{000000} \textbf{Image 3}}}  & great                                                                & \cellcolor[HTML]{34FF34}great                                    & \cellcolor[HTML]{FFC702}good                                      & \cellcolor[HTML]{FFC702}good                                   \\ \hline
\multicolumn{1}{|c|}{\cellcolor[HTML]{FFFFFF}{\color[HTML]{000000} \textbf{Image 4}}}  & great                                                                & \cellcolor[HTML]{FFC702}good                                     & \cellcolor[HTML]{FFC702}good                                      & great                                                          \\ \hline
\multicolumn{1}{|c|}{\cellcolor[HTML]{FFFFFF}{\color[HTML]{000000} \textbf{Image 5}}}  & \cellcolor[HTML]{FFC702}good                                         & \cellcolor[HTML]{FFC702}good                                     & \cellcolor[HTML]{34FF34}great                                     & great                                                          \\ \hline
\multicolumn{1}{|c|}{\cellcolor[HTML]{FFFFFF}{\color[HTML]{000000} \textbf{Image 6}}}  & great                                                                & \cellcolor[HTML]{FD6864}bad                                      & \cellcolor[HTML]{34FF34}great                                     & great                                                          \\ \hline
\multicolumn{1}{|c|}{\cellcolor[HTML]{FFFFFF}{\color[HTML]{000000} \textbf{Image 7}}}  & great                                                                & \cellcolor[HTML]{FD6864}bad                                      & \cellcolor[HTML]{34FF34}great                                     & great                                                          \\ \hline
\multicolumn{1}{|c|}{\cellcolor[HTML]{FFFFFF}{\color[HTML]{000000} \textbf{Image 8}}}  & great                                                                & \cellcolor[HTML]{FD6864}bad                                      & \cellcolor[HTML]{FD6864}bad                                       & great                                                          \\ \hline
\multicolumn{1}{|c|}{\cellcolor[HTML]{FFFFFF}{\color[HTML]{000000} \textbf{Image 9}}}  & great                                                                & \cellcolor[HTML]{FD6864}bad                                      & \cellcolor[HTML]{34FF34}great                                     & great                                                          \\ \hline
\multicolumn{1}{|c|}{\cellcolor[HTML]{FFFFFF}{\color[HTML]{000000} \textbf{Image 10}}} & great                                                                & \cellcolor[HTML]{34FF34}great                                    & \cellcolor[HTML]{FD6864}bad                                       & great                                                          \\ \hline

\multicolumn{1}{|c|}{\cellcolor[HTML]{FFFFFF}{\color[HTML]{000000} \textbf{Image 11}}} & \cellcolor[HTML]{FFC702}good &  \cellcolor[HTML]{FD6864}bad & \cellcolor[HTML]{34FF34}great & \cellcolor[HTML]{FFC702}good \\ \hline

\multicolumn{1}{|c|}{\cellcolor[HTML]{FFFFFF}{\color[HTML]{000000} \textbf{Image 12}}} & \cellcolor[HTML]{FFC702}good &  \cellcolor[HTML]{FFC702}good & \cellcolor[HTML]{FD6864}bad & \cellcolor[HTML]{34FF34}great \\ \hline
 
\end{tabular}
\end{table}

As a way of carrying out an experiment in an equal way, precision was used in the recognition of each class in the images in the applications. Each application was classified according to its prediction, with 3 statuses being assigned, according to the result obtained: Bad (if the app detetcs less than 50\% of the food in the image), Good (if it detects more than 50\% of the food) and Great (if it detects all foods in the image). In this sample and experiment the results showed that MyFood obtained better prediction when compared to existing solutions.

% Uma análise preditiva sobre novos dados foi realizada a fim de testar os modelos implementados em comparação outras aplicações de mercado, para identificar qual possui o melhor desempenho em relação ao processo de segmentação e classificação de alimentos.

\section{Conclusion and Future Work}

This research focused on the development of a computer system for food recognition, through image classification and segmentation. The proposed system is capable of recognizing 9 classes of food, using deep learning methods and image segmentation algorithms. As main contribution, we built a Brazilian food dataset and conducted experiments of SOTA segmentation methods to perform food segmentation.

The segmentation analysis showed that FCN and Mask-RCNN obtained better results of segmentation, with a value of IoU of 0.70. The prototype developed in this research, using Mask-RCNN obtained better results compared with other approaches used in the market. 

%Demonstrating that the segmentation algorithms, in addition to being more specialized in the classification of each item in the image, also proved to be effective in the integration and recognition of foods.

% Image segmentation, despite being a recent challenge for deep neural networks, was shown through its architectures that have a great potential for understanding and understanding the global visual context of the entrance, which significantly reflects its prediction. The architectures link different parts of the image to understand the relationships between objects, and thus be able to have multiple classifications of objects with a significant degree of recognition. Thus, it is concluded that the methodology implemented in these models, for classification and segmentation of food images, proved to be a good approach for the recognition of patterns in food images.

% As a proposal for improvements and future work, the parameters optimization process of the models will be carried out so that they reach a more efficient classification and segmentation.
In future work, we will add 2D image volume analysis methods, and we will explore new methods to calculate the volume of food more accurately and thus have more nutritional information.

% conference papers do not normally have an appendix

% use section* for acknowledgment
% \section*{Acknowledgment}

% The authors would like to thank...

% trigger a \newpage just before the given reference
% number - used to balance the columns on the last page
% adjust value as needed - may need to be readjusted if
% the document is modified later
%\IEEEtriggeratref{8}
% The "triggered" command can be changed if desired:
%\IEEEtriggercmd{\enlargethispage{-5in}}

% references section

% can use a bibliography generated by BibTeX as a .bbl file
% BibTeX documentation can be easily obtained at:
% http://mirror.ctan.org/biblio/bibtex/contrib/doc/
% The IEEEtran BibTeX style support page is at:
% http://www.michaelshell.org/tex/ieeetran/bibtex/
\bibliographystyle{IEEEtran}
% argument is your BibTeX string definitions and bibliography database(s)
\bibliography{example}
%
% <OR> manually copy in the resultant .bbl file
% set second argument of \begin to the number of references
% (used to reserve space for the reference number labels box)
%\begin{thebibliography}{1}
%
%\bibitem{IEEEhowto:kopka}
%H.~Kopka and P.~W. Daly, \emph{A Guide to \LaTeX}, 3rd~ed.\hskip 1em plus
%  0.5em minus 0.4em\relax Harlow, England: Addison-Wesley, 1999.

%\end{thebibliography}

% that's all folks
\end{document}